*Article*

# Entropy Approximation by Machine Learning Regression: Application for Irregularity Evaluation of Images in Remote Sensing

**Andrei Velichko** [1,*]**, Maksim Belyaev** [1]**, Matthias P. Wagner** [2] **and Alireza Taravat** [3]

[1] Institute of Physics and Technology, Petrozavodsk State University, 185910 Petrozavodsk, Russia; biomax89@yandex.ru

[2] Panopterra, 64293 Darmstadt, Germany; matthias.wagner@panopterra.com

[3] Deimos Space, Oxford OX110QR, UK; art23130@gmail.com

[*] Correspondence: velichko@petrsu.ru; Tel.:+7-911-400-5773

**Abstract:** Approximation of entropies of various types using machine learning (ML) regression methods are shown for the first time. The ML models presented in this study define the complexity of the short time series by approximating dissimilar entropy techniques such as Singular value decomposition entropy (SvdEn), Permutation entropy (PermEn), Sample entropy (SampEn) and Neural Network entropy (NNetEn) and their 2D analogies. A new method for calculating $SvdEn_{2D}$, $PermEn_{2D}$ and $SampEn_{2D}$ for 2D images was tested using the technique of circular kernels. Training and testing datasets on the basis of Sentinel-2 images are presented (two training images and one hundred and ninety-eight testing images). The results of entropy approximation are demonstrated using the example of calculating the 2D entropy of Sentinel-2 images and $R^2$ metric evaluation. The applicability of the method for the short time series with a length from $N = 5$ to $N = 113$ elements is shown. A tendency for the $R^2$ metric to decrease with an increase in the length of the time series was found. For SvdEn entropy, the regression accuracy is $R^2 > 0.99$ for $N = 5$ and $R^2 > 0.82$ for $N = 113$. The best metrics were observed for the $ML\_SvdEn_{2D}$ and $ML\_NNetEn_{2D}$ models. The results of the study can be used for fundamental research of entropy approximations of various types using ML regression, as well as for accelerating entropy calculations in remote sensing. The versatility of the model is shown on a synthetic chaotic time series using Planck map and logistic map.

## 1. Introduction

Entropy is a measure of chaos or irregularity. It is also a measure of the structural organization of systems since order and chaos are not only opposite but also complementary concepts. Entropy as a measure of chaos or order was described for systems of a different nature: the Clausius entropy in thermodynamics [1], the Boltzmann entropy [2] in statistical physics, the Shannon entropy [3] in information theory, the Kolmogorov entropy [4] in the theory of dynamical systems, the von Neumann entropy in quantum mechanics [5]. Understanding the universality of entropy as a measure of chaos, regardless of the nature of the system, came gradually, starting with the works of Kolmogorov [4], Renyi [6], Shannon [3] and others, and it continues in the present day [7]. In this paper, an attempt was made to develop a methodology for a universal approach to estimate entropy using machine learning (ML) technologies. The results of the approximation of the entropies of various types using ML regression are presented.

In real-world systems, not only spatial, but also temporal structures are found. Temporal structures are usually studied by analyzing the time series of observational data; spatial structures, as a rule, are usually studied by analyzing two-dimensional images. Along with time series, images are the most important sources of information in remote sensing (RS) imagery [8] and geophysical mapping [9]. Different versions of entropies have been used for remote sensing applications. For example, Shannon's entropy has been used for measuring urban sprawl [10,11], and exponential entropy has been used for image segmentation [12]. The most common entropy way of obtaining entropy, however, is still the use of first- and second-order texture metrics based on quantized Gray-Level Co-occurrence Matrices (GLCM) [13]. Entropy is also used in image quality assessments [14,15] and change detection [16]. Two-dimensional dispersion entropy (DispEn$_{2D}$) [17], sample entropy (SampEn$_{2D}$) [18], permutation entropy (PerEn$_{2D}$) [19], approximate entropy (ApEn$_{2D}$) [20] and Neural Network entropy (NNetEn$_{2D}$) [21] have been proposed for image processing applications, and they can be considered as an irregularity measure for images. To calculate 2D entropy, the approach of transforming a 2D kernel into a 1D time series is often used. For example, a square kernel with alternating row and column readouts could be used [18]; in addition, the transformation method using the Hilbert–Peano curve is popular due to its low computational cost and its ability to preserve the relevant properties of pixel spatial correlation [22–24]. In this work, we applied our own method of transforming a 2D image area into a 1D time series using a circular kernel; the advantage of this is the stability of the result after image rotation [21].

Regression is a method of investigating the relationship between the independent variables or features and an outcome. It is used as a predictive modeling technique in machine learning in which an algorithm is used to predict continuous results. There are many ML algorithms for regression analysis: gradient boosting (GB), support vector regression (SVR), k-nearest neighbors (KNN), multi-layer perceptron (MLP), stochastic gradient descent (SGD), decision tree (DT), automatic relevance determination (ARD), adaptive boosting (AB), etc.

In this paper, we have developed a technique for approximating several types of entropy using ML regression: Singular value decomposition entropy, Permutation entropy, Sample entropy and Neural Network entropy. The block diagram of the two approaches to calculating the entropy is shown in Figure 1. The input data are the time series, and the output is the value of the entropy of the time series. In the first approach, the entropy can be calculated by the use of standard techniques, for example, using well-known algorithms for calculating SvdEn, SampEn, PermEn and NNetEn. In this work, we have shown the existence of another approach in which each type of entropy can be approximated by ML regression which is trained on the training dataset. Each regression model was trained on time series of a certain length and for a certain type of entropy. The entropies that were calculated using ML regression are denoted by the prefix 'ML'. The paper presents training and testing datasets. The application to remote sensing is shown by calculating the 2D entropy distribution of Sentinel-2 images, and $R^2$ estimates of the approximation error were made. It is shown that the best model of ML regression is the gradient boosting algorithm, in which the standard entropy practically coincides with the result of ML regression. The performance of the entropy approximation model on short time series up to $N = 113$ elements is shown. The versatility of the model is shown on a synthetic chaotic time series using Planck map and logistic map.

In addition, the use of circular kernels for transforming images into a series of time series and calculating the 2D distribution of SvdEn (SvdEn$_{2D}$), SampEn (SampEn$_{2D}$) and PermEn (PermEn$_{2D}$) is introduced. We used a similar approach earlier in [21] to calculate the 2D distribution of NNetEn (NNetEn$_{2D}$).

The paper has the following structure. Section 2 describes the methods, dataset preparation and the training process. Section 3 presents the results of the entropy estimation, which are followed by a discussion in Section 4. Section 5 summarizes the main conclusions.

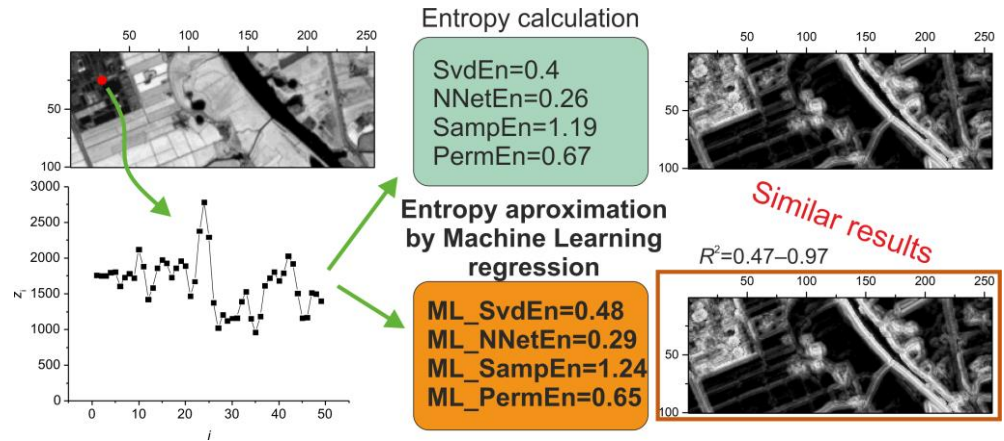

**Figure 1.** Block diagram of two approaches to entropy calculation: the standard methods and the method of entropy approximation via ML regression models ('ML_'). The 2D kernel (visualized in red color in the top left image) is transformed to a 1D time series for entropy approximation.

## 2. Methods and Datasets

### 2.1. Time Series Normalization Method

The main idea of normalization is to bring the time series to a standard form before the calculation or passing it to the ML model for a subsequent entropy approximation.

Let us denote the set of elements of the time series as $Z = [z_1, z_2, \dots z_i, \dots z_N]$, where $z_i$ is the element of the set ($i = 1 \dots N$), and $N$ is the number of elements of the time series. The maximum and minimum values of the series are denoted as:

$$z_{\min} = \min(Z)$$
$$z_{\max} = \max(Z) \tag{1}$$

Let us denote as $T_{upp}$—the upper threshold of normalization; $T_{low}$—the lower threshold of normalization; these were calculated using the formulas.

$$if \ \ z_{\min} = z_{\max} \ \text{then} \ T_{low} = 0, \ T_{upp} = 0 \tag{2}$$

$$if \ \ |z_{\min}| \geq |z_{\max}| \ \text{then} \ \ T_{low} = -|z_{\min}|, T_{upp} = z_{\max} + EN \cdot (|z_{\min}| - z_{\max}) \tag{3}$$

$$if \ \ |z_{\min}| < |z_{\max}| \ \text{then} \ T_{low} = z_{\min} - EN \cdot (|z_{\max}| + z_{\min}), \ T_{upp} = |z_{\max}| \tag{4}$$

The $EN$ value is called the normalization parameter, and it varies from 0 to 1.

$$0 \leq EN \leq 1 \tag{5}$$

The final step is the normalization of the time series $Z$ to the series $X = [x_1, x_2, \dots x_i, \dots x_N]$, according to the formula:

$$\begin{cases} if \ \ (T_{upp} - T_{low}) \neq 0 \ \text{then} \ x_i = \dfrac{(z_i - T_{low})}{T_{upp} - T_{low}} \cdot 2 - 1 \\ if \ \ (T_{upp} - T_{low}) = 0 \ \text{then} \ x_i = 0 \end{cases} \tag{6}$$

The normalization parameter $EN$ allows us to smoothly change the degree of filtering of the constant component of the time series. An example of the effect of $EN$ is shown in Figure 2, where the initial time series $Z$ containing $N = 49$ elements (Figure 2a) is normalized into time series $X$ (Figure 2b). If $EN = 1$, then normalization does not remove the constant component of the time series. When $EN = 0$, normalization translates the value range of the time series into an interval from −1 to 1, and the constant component is suppressed (Figure 2b). A value of $EN = 0.5$ results in a partial suppression of the DC component.

Thus, the choice of the parameter $EN$ allows, in the course of normalization, us to filter the variable component of the time series at $EN = 0$. The $EN$ parameter is of great importance when the entropy depends on the DC component, in which case the $EN$ value significantly affects the result of the 2D entropy calculation.

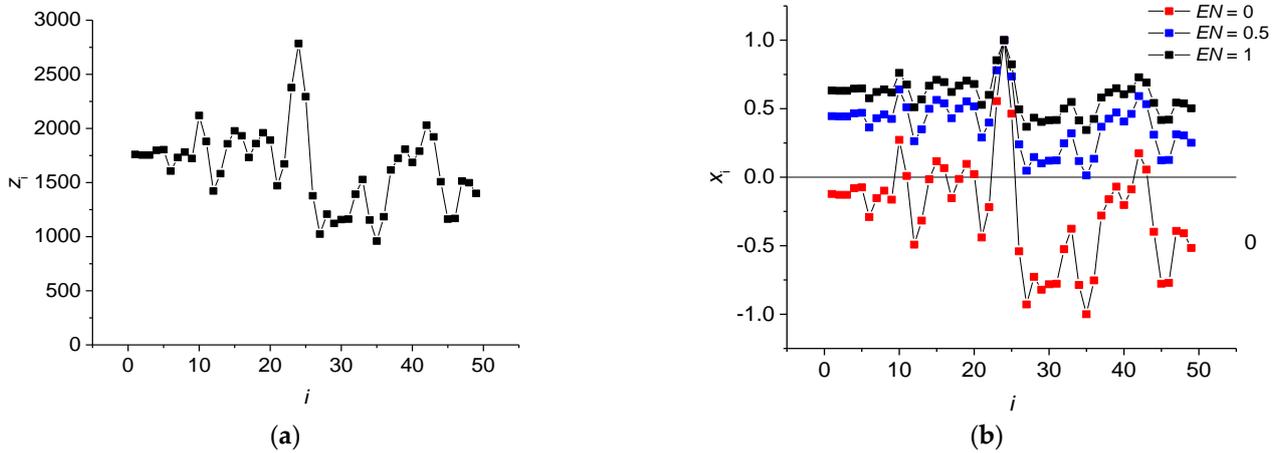

(**a**)                                                                         (**b**)

**Figure 2.** An example of the time series $Z$ ($N = 49$) (**a**) and the result of its normalization to the $X$ series for $EN = 0$, $EN = 0.5$ and $EN = 1$ (**b**).

## 2.2. Methods for Entropy Evaluation with Standard Methods

In this article, several types of entropies were calculated: Singular value decomposition entropy, Permutation entropy, Sample entropy and Neural Network entropy, the basic principles of which are disclosed below. For the calculations, the open-source python library 'antropy' was used [25].

### 2.2.1. Singular Value Decomposition Entropy

Singular value decomposition (SVD) is an analogue of the spectral decomposition of the signals, which is applicable to arbitrary matrices. This transformation was first proposed by E. Beltrami in 1873 [26]. SVD is the factorization of a matrix $A$ into the product.

$$A = USV^T \tag{7}$$

The matrix $U$ contains the left singular vectors of $A$, and the matrix $V$ contains the right singular vectors. The matrix $S$ is always diagonal, its coefficients are non-negative real numbers $\lambda_1, ..., \lambda_k$, which are located on the main diagonal of the matrix, and they are called singular values.

For a time series $X = [x_1, x_2, ... x_i, ... x_N]$, $d$ dimension phase space can be reconstructed by means of sliding with window $a(i)$. The procedure is as follows.

$$a(i) = [x_i, x_{i+delay}, ..., x_{i+(d-1) \cdot delay}]$$
$$A = [a(1), a(2), ..., a(N - (d-1) \cdot delay)]^T \tag{8}$$

where $d$—the length of the embedding dimension; $delay$—the time series sample bias.

The dispersion of the singular values $\lambda_k$ also provides an indication of the complexity of the signal dynamics [27]. Singular values can be normalized as:

$$\overline{\lambda}_k = \frac{\lambda_k}{\sum \lambda_k} \tag{9}$$

Singular value decomposition entropy is defined with the Shannon formula which was applied to the elements of singular values of the matrix, and this was calculated as follows [27]:

$$\text{SvdEn} = -\sum \overline{\lambda}_k \cdot \ln \overline{\lambda}_k \tag{10}$$

After that, the SvdEn values were normalized in the range from 0 to 1:

$$\text{SvdEn} = \frac{\text{SvdEn}}{\log_2 d} \tag{11}$$

SvdEn is used in a number of works: to analyze the heart rate variability [27], to analyze effect of taping on ankle joint dynamics [28], as an indicator of the state of financial markets [29,30], for a quantification of ecological complexity [31] and in computer graphics [32].

Image processing was carried out with the length of the embedding dimension $d = 3$ and $delay = 1$.

### 2.2.2. Permutation Entropy

The permutation entropy is a complexity measure for the time series based on the comparison of neighboring values. The permutation entropy PermEn of a one-dimensional data series $X$ is:

$$\text{PermEn} = -\sum p_i \cdot \log_2 p_i \tag{12}$$

where $p_i$ is the frequency of the occurrence of the $i$-th permutation in the embedded matrix $A$, which is defined in the same way as in (8).

After that, the PermEn values were normalized in the range from 0 to 1:

$$\text{PermEn} = \frac{\text{PermEn}}{\log_2 d!} \tag{13}$$

Image processing was carried out with the length of the embedding dimension $d = 5$ and $delay = 5$.

### 2.2.3. Sample Entropy

The Sample entropy calculation of the time series $X = [x_1, x_2, \dots x_N]$ of the length contains several stages. First, the series was divided into template vector $X_m(i) = [x_i, x_{i+1}, \dots x_{i+m-1}]$ of length $m$ ($m < N$). Then, the number of template vectors $X_m$ were counted, and the Chebyshev distance between them $d[X_m(i), X_m(j)]$ ($i \neq j$) does not exceed $r$. The sample entropy is a variant of the Approximation Entropy.

The sample entropy for the one-dimensional data series $X$ is defined as:

$$\text{SampEn} = -\ln(\frac{C(m+1, r)}{C(m, r)}) \tag{14}$$

where $C(m,r)$ is the number of pairs of vectors of length $m$, the distance between which does not exceed $r$.

In this work, the image processing was carried out at $m = 2$ and $r = 0.2\sigma$, where $\sigma$ —is the standard deviation within the time series $X$.

### 2.2.4. Neural Network Entropy

Neural network entropy is the first entropy measure that is based on artificial intelligence methods, and it was introduced in [33] for 1D time series (NNetEn₁D), and then, it was extended to calculate the entropy of 2D images (NNetEn₂D) [21]. It computes entropy di-rectly, without considering or approximating probability distributions. NNetEn is computed using the LogNNet neural network. The LogNNet model [34] was originally designed for recognizing handwritten digits in the MNIST dataset [35] with 60,000 images for training and 10,000 images for testing. It comprises three parts (see Figure 3): the input layer, a model reservoir of matrix $W_1$ to transform the input vector $Y$ into an intermediate vector, and a single layer feedforward neural network transforming vector $S_h$ into digits 0–9 in the output layer $S_{out}$.

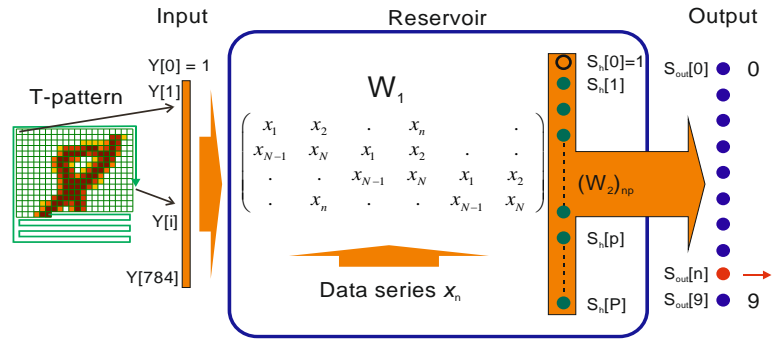

**Figure 3.** The LogNNet model structure for NNetEn calculation [33].

To determine NNetEn, the LogNNet reservoir matrix was filled with elements of the studied data $X = [x_1, x_2, \ldots x_N]$. The network was then trained and tested on MNIST-10 datasets to obtain classification accuracy. This accuracy is considered as the entropy measure, and it is denoted as NNetEn$_{1D}$.

$$\text{NNetEn} = \frac{\text{Classification accuracy}}{100\%} \tag{15}$$

The procedure for calculating NNetEn$_{1D}$ is described in more detail in [33].

LogNNet can be used for estimating the entropy of the time series as the transformation of inputs is carried out by the time series, and this affects the classification accuracy. A more complex transformation of the input information, which is performed by the time series in the reservoir part, results in a higher classification accuracy in LogNNet.

The maximum length of the time series that can be fed to the model is determined by the number of elements in matrix $W_1$ ($N_0 = 19625$). The main technique used in this work for filling the matrix was the W1M_1 method (filling by rows as in Figure 3 with copying of the series) and the epoch number $Ep = 4$ [21].

### 2.2.5. Method for 2D Entropy Calculation with Circular Kernels

The method of using circular kernels to calculate 2D entropy was developed in [21]. Its essence lies in the transformation of a 2D image area using circular kernels of a radius $R$ into a one-dimensional series. Figure 4a shows the principle of covering the entire image with circular kernels using a step $S$ and an initial offset $DL$. Areas that are outside the image boundaries are not defined, so the pixel values in these areas are filled by the symmetrical mirroring of the pixels in the image. To calculate NNetEn$_{2D}$, the set of pixels inside the local kernel was converted into a one-dimensional data series (Figure 4b), and then, it was calculated in the same way as in the one-dimensional case. In Figure 4b, the sequence of the formation of the elements of the series can be traced along the connecting red line, starting from the center of the kernel $n = 1$ and ending with element $n = 49$. The number of pixels $N$ in a circular kernel has a quadratic dependence on the radius, and the evaluation of the sample values is given in Table 1. Example kernels for $R = 6$ and $R = 1$ are shown in Figure 5a,b. In this research, we used $DL = 0$ and step $S = 1$.

**Table 1.** Number of pixels $N$ in a circular kernel versus radius $R$.

| $R$ | 1 | 2 | 3 | 4 | 5 | 6 |
|---|---|---|---|---|---|---|
| $N$ | 5 | 13 | 29 | 49 | 81 | 113 |

The entropy distribution on a 2D image is denoted as: $SvdEn_{2D}$, $SampEn_{2D}$, $PermEn_{2D}$ and $NNetEn_{2D}$.

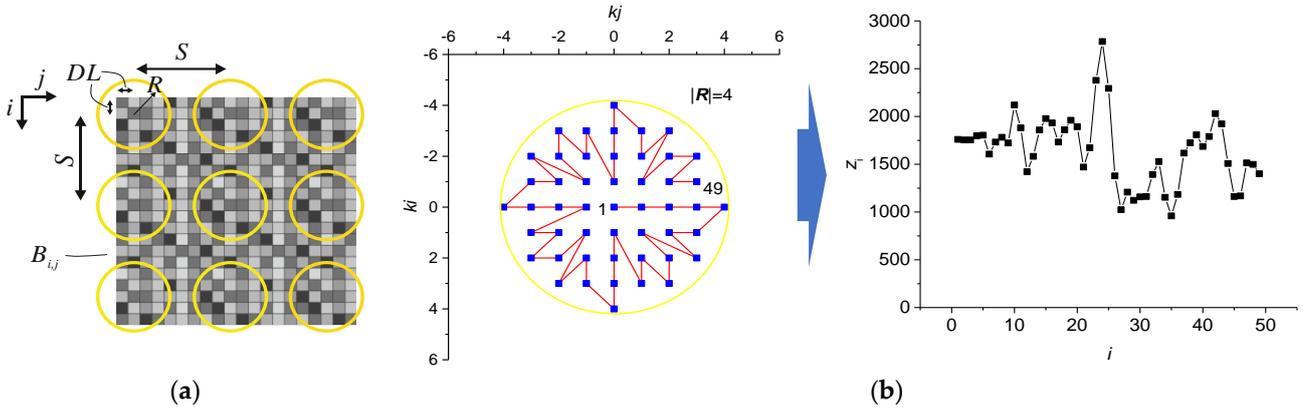

**Figure 4.** (**a**) Scheme of filling the image with circular kernels for 2D entropy calculation. (**b**) Scheme for converting a two-dimensional pixel distribution into a one-dimensional data series $z_i$ ($R = 4$).

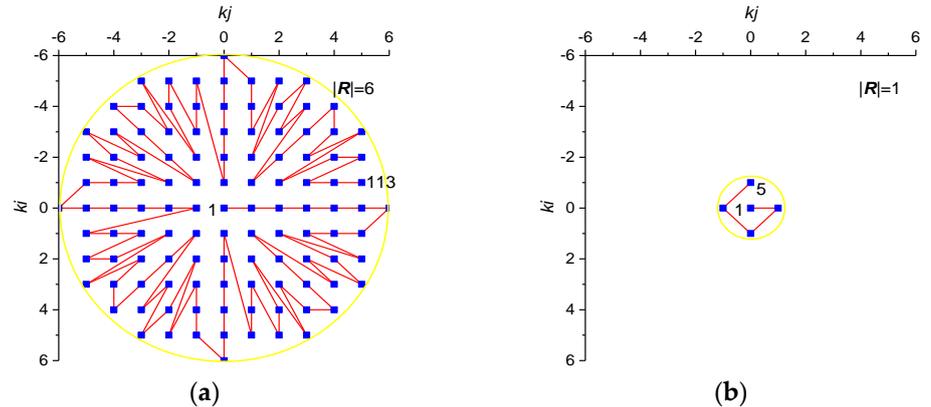

**Figure 5.** Scheme for converting a two-dimensional pixel distribution into a one-dimensional data series with different radius of circular kernel (**a**) $R = 6$; (**b**) $R = 1$.

### 2.3. Entropy Approximation by ML Regression Models

To approximate the entropy, a regression method based on machine learning was used. We will denote the approximation of the entropies which were calculated using the ML method as ML_"Entropy method".

The overall plan for training and testing consisted of the following main steps:

1. Setting the entropy type. We used 4 types of entropy: SvdEn, SampEn, PermEn and NNetEn. We denoted their approximations using ML regression as ML_SvdEn, ML_SampEn, ML_PermEn and ML_NNetEn, respectively.
2. Setting the ML algorithm for regression: a gradient boosting algorithm was used as the main method.
3. Setting the length of the time series $N$. In this research, we tested several lengths of short time series $N$ = 5, 13, 29, 49, 81 and 113 (see Table 1).
4. Generation of training dataset using two images. Each element of the training set consisted of a time series of length $N$ and an output entropy value.
5. Hyperparameter optimization and training of the regression model using training dataset.

6.  Generation of a test dataset based on 198 sample images from Sentinel-2. One element of the test set consisted of a time series of length $N$ and the output entropy value.
7.  Testing the regression model on a test dataset and determining the error using $R^2$ metric. At the input of the ML algorithm, it is necessary to supply a vector of a time series of a certain length on which the algorithm was trained.
8.  Calculation of 2D entropies using circular kernels: $SvdEn_{2D}$, $SampEn_{2D}$, $PermEn_{2D}$ and $NNetEn_{2D}$.

    Below is a more detailed description of some of these steps.

### 2.3.1. Dataset Description

To train and test the entropy approximation models, we used a dataset of 200 Sentinel-2 sample images (see Supplementary Materials). Each image was of size 256 × 256, and they contained four bands (blue, green, red and near-infrared). To identify the images, we used the number $n$ in the dataset from $n$ = No. 1 to No. 200.

Example images for the near-infrared band are shown in Figure 6.

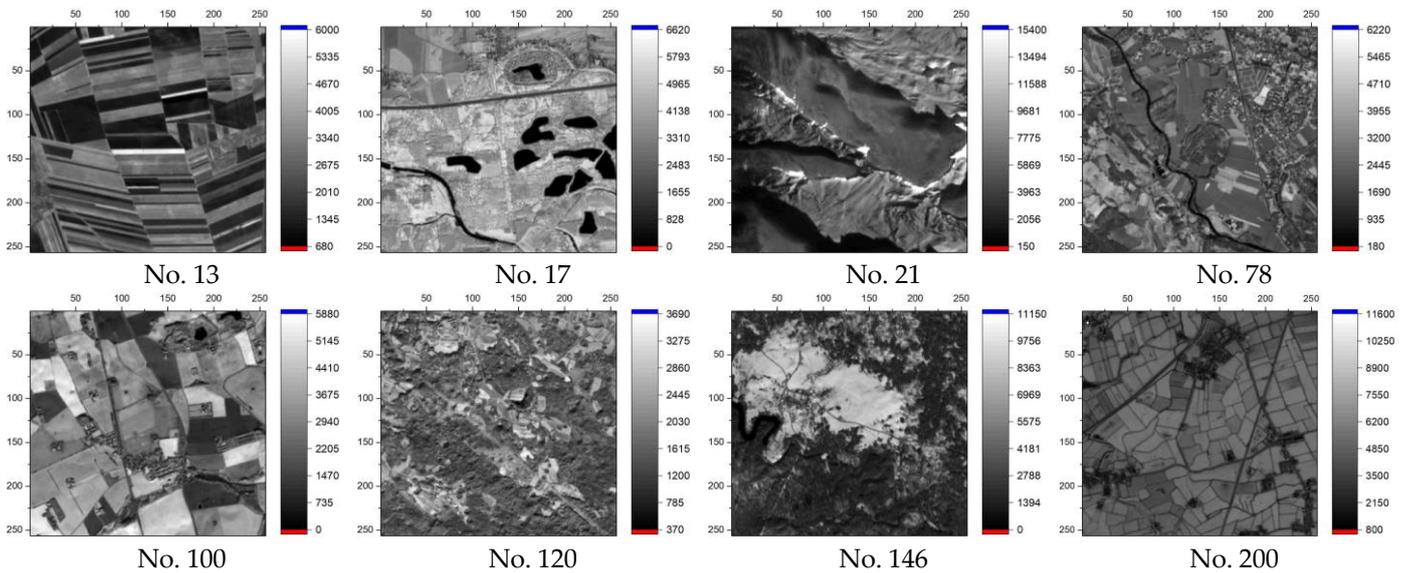

**Figure 6**. Sample images from the Sentinel-2 images database (near-infrared band).

### 2.3.2. Training Dataset

For each type of entropy ML_SvdEn, ML_PermEn, ML_SampEn and NNetEn, as well as a given length of the time series $N$, a separate training set was generated.

To generate the training set, we used two images numbered No. 1 and No. 2, which are shown in Figure 7 (near-infrared band). One element of the set consisted of a time series of length $N$ and the entropy value. Time series were generated at each pixel using a circular kernel as shown in Section 2.2.5. For two images, the number of pixels 256 × 256 × 2 = 131,072. Before calculating the entropy, each time series was normalized according to the method in Section 2.1. In addition, the entropy was calculated for two normalization options for $EN$ = 0 and $EN$ = 1. As a result, the total number of elements in the training dataset was 131,072 × 2 = 262,144.

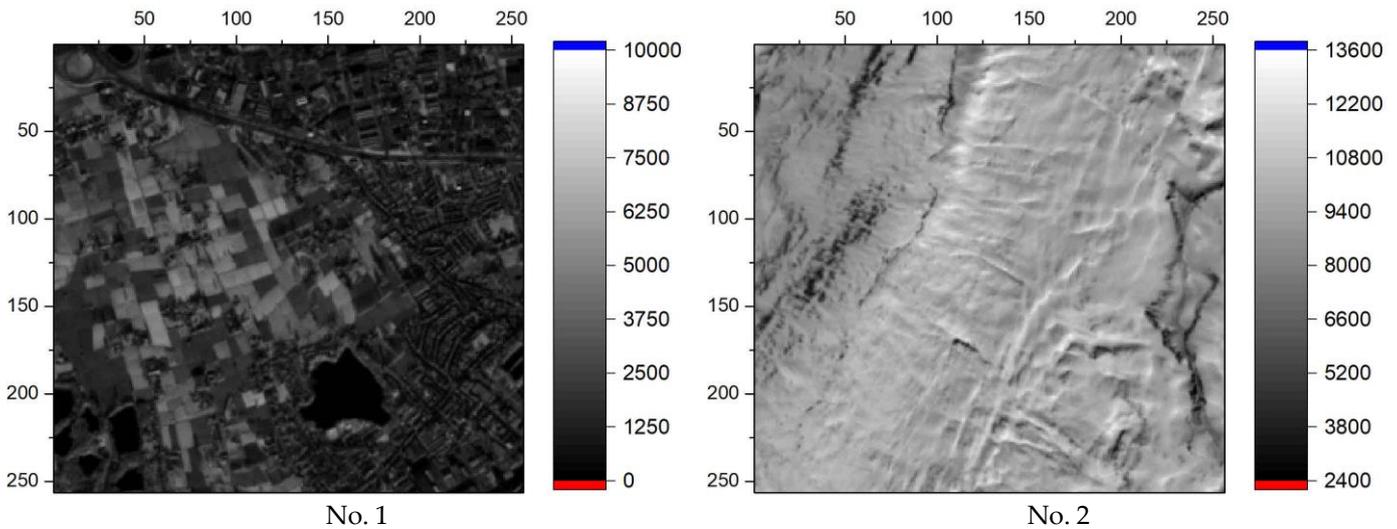

**Figure 7.** Images No. 1 and No. 2 from the database that were used to train ML regression models (near-infrared band).

### 2.3.3. Hyperparameter Optimization and Training the Regression Model

The package "auto-sklearn" library [36] was used for hyperparameter optimizations and training of the regression models. This library allows you to select the most accurate models that are obtained with different values of hyperparameters.

The regression models were trained using K-fold cross-validation. In this case, the training set was divided into $K$ parts, and the model was trained $K$ times. Moreover, at each iteration of training, various combinations of $K$-1 parts of the training dataset were used. After the training, the model was validated on the part of the dataset that was not used during training, and the model error metric $R^2$ was calculated (see Section 2.3.5). After performing $K$ validations, the obtained $R^2$ values were averaged, and this average $R^2(1,2)$ value was used to assess the accuracy of the model (indices in brackets mean that the metric was calculated on the set of two images No. 1 and No. 2). This approach made it possible to obtain an adequate assessment of the accuracy of the model and minimize the effects of random sampling for testing and validation.

### 2.3.4. Test Dataset Generation

The test dataset for ML_SvdEn, ML_PermEn and ML_SampEn was generated based on 198 images in the database ($n = 3$–200). Each pixel of each image represents one row of the dataset (time series and entropy), so there were $256 \times 256 = 65,536$ rows per picture and $65,536 \times 198 = 12,976,128$ rows in total. The test dataset for ML_NNetEn was generated in a similar way, except that a smaller number of pixels per image (1936) was used since NNetEn calculations take a long time. Before calculating the entropy, each time series was normalized according to the method described in Section 2.1.

For ML_SvdEn, the test datasets were separately generated for four bands (blue, green, red and near-infrared) with $EN$ values from 0 to 1 with a step of 0.1. For ML_NNetEn, the test sets were generated for near-infrared band with three values of $EN = 0$, 0.5 and 1. For ML_PermEn and ML_SampEn, the test sets were generated only for $EN = 1$ since these entropies do not depend on $EN$.

The test sets ML_SvdEn and ML_NNetEn contained elements with variable $EN$ that are not included in the training set (for example, $EN = 0.5$). This was conducted to assess the universality of the approximation technique. One of the tasks pursued in the work is the creation of a universal method for entropy approximation for the entire range of $EN$ values (Equation (5)).

In addition, testing in some cases was also carried out on red, green and blue bands, which also tests the versatility of the technique.

It can be seen that the test dataset significantly exceeds the training set in terms of the number of elements since it was tested on a larger number of images (2 vs. 198), a wider range of *EN*, as well as on red, green and blue bands.

### 2.3.5. Estimation the Accuracy of the ML Model

To assess the accuracy of the regression models, we used the coefficient of determination (or R² metric), which was determined for each picture from the test set, separately. To calculate R², the values of the entropies $Y = [y_1, y_2, …, y_{NN}]$ and the approximations of the entropies $MLY = [MLy_1, MLy_2, …, MLy_{NN}]$ which were calculated by the regression models on time series which was obtained within the same picture were used. The *NN* value is the size of the set within one picture, which is equivalent to the number of pixels ($NN = 256 \times 256 = 65,536$). First, the average value $y'$ was calculated for a set of entropy values $y$:

$$y' = \frac{1}{NN} \sum_{i=1}^{NN} y_i \tag{16}$$

Then, sum of squares of residuals *SS*res and total sum of squares *SS*tot were calculated:

$$SS_{res} = \sum_{i=1}^{NN} (y_i - MLy_i)^2 \tag{17}$$

$$SS_{res} = \sum_{i=1}^{NN} (y_i - y')^2 \tag{18}$$

Using *SS*res and *SS*tot, R² was calculated as follows:

$$R^2 = 1 - \frac{SS_{res}}{SS_{tot}} \tag{19}$$

If all of the pairs of values $y_i$ and $MLy_i$ match, the value of *SS*res will be equal to 0 and R² will be equal to 1, in other cases R² < 1. Thus, the closer R² is to 1, the more accurately the regression model describes the test data.

The number of the image *n* for which the calculation is being performed is indicated in brackets R²(*n*). R²(1,2) means that the metric was calculated on the set of two images No. 1 and No. 2 from the training dataset.

The Pearson correlation coefficient was used to assess the accuracy of the synthetic data.

### 2.3.6. Synthetic Time Series Approximation Method

To show the versatility, efficiency and robustness of the proposed method of entropy approximation, we apply the method to synthetic time series $x_{1…xN}$ with a length $N = 29$. The time series generated by the chaotic Planck map (Equation (20), Figure 8) was chosen as a training sample. Because of the transient period, the first 1000 elements were ignored.

Planck map:

$$x_{n+1} = \frac{r \cdot x_n^3}{1 + \exp(x_n)}, \ 3 \le r \le 7, \ x_{-999} = 4. \tag{20}$$

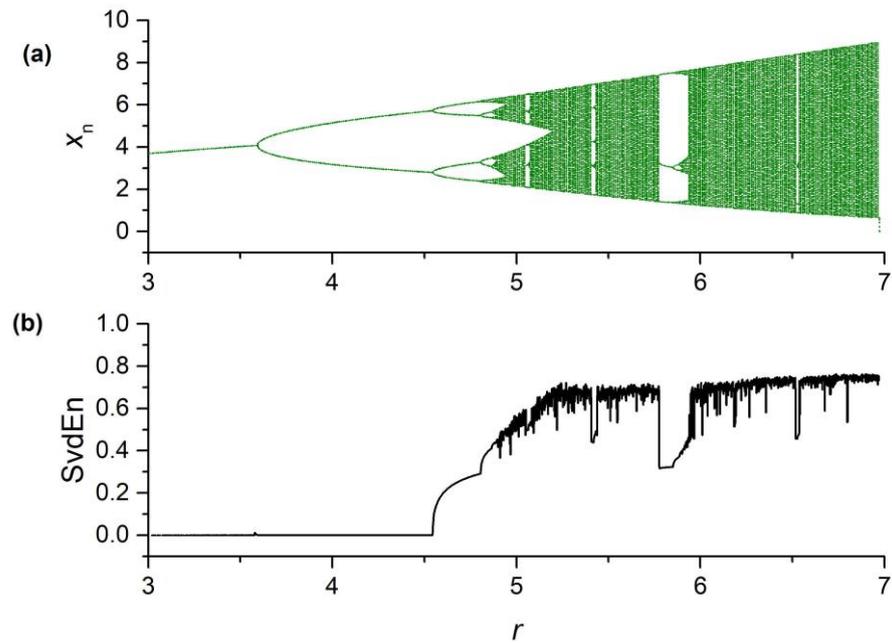

**Figure 8.** Bifurcation diagrams for Planck map (Equation (20)) (**a**); the dependence of SvdEn on the parameter $r$ ($N = 29$, $EN = 0$, $d = 20$ and *delay* = 1) (**b**).

The training synthetic dataset consisted of 100,000 time series, each of a length $N = 29$, which were generated over the entire range of $r$. Figure 8a shows the bifurcation diagrams for the Planck map and the corresponding SvdEn values in Figure 8b.

The synthetic testing datasets were a set of 3000 time series which were generated by logistic mapping (Equation (21)).

Logistic map:

$$x_{n+1} = r \cdot x_n \cdot (1 - x_n)$$, $1 \le r \le 4$, $x_{-999} = 0.1$. (21)

Thus, the training and testing synthetic datasets had different time series generation algorithms.

The calculation of SvdEn for the synthetic data was carried out with the length of the embedding dimension $d = 20$ and *delay* = 1. The normalization was preliminarily performed according to Equations (1)–(6) with the parameter $EN = 0$.

## 3. Results

### 3.1. Calculation of 2D Entropy with Variation of the Normalization Parameter EN

Let us take image No. 172 (Figure 9) as an example for calculating 2D entropy. The kernel radius was $R = 4$, and its shape and size are visualized in red. The 2D entropy distributions (SvdEn2D and NNetEn2D) for the normalization parameters $EN = 0$, $EN = 0.5$ and $EN = 1$ are shown in Figure 10. It can be seen that for SvdEn2D and NNetEn2D, the entropy value depends on the parameter $EN$.

The 2D entropy distributions (SampEn2D and PermEn2D) for the normalization parameters $EN = 0$, $EN = 0.5$ and $EN = 1$ are shown in Figure 11. It can be seen that for SampEn2D and PermEn2D, the entropy value does not depend on the parameter $EN$. These entropies are sensitive only to the variable component of the time series and do not depend on the constant component.

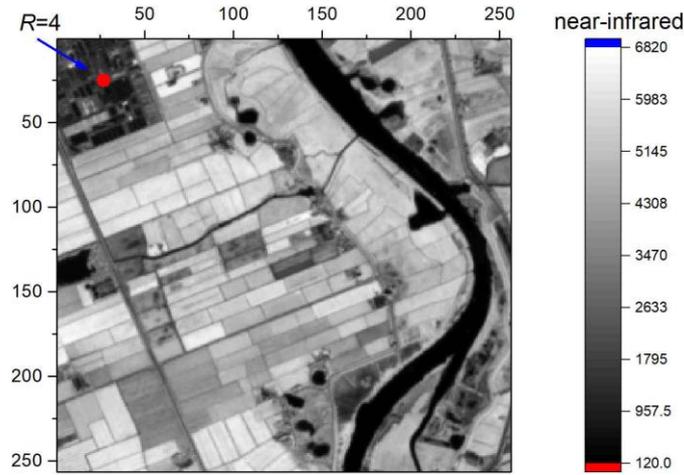

**Figure 9.** Image No. 172, with 2D kernel visualization in red color ($R = 4$).

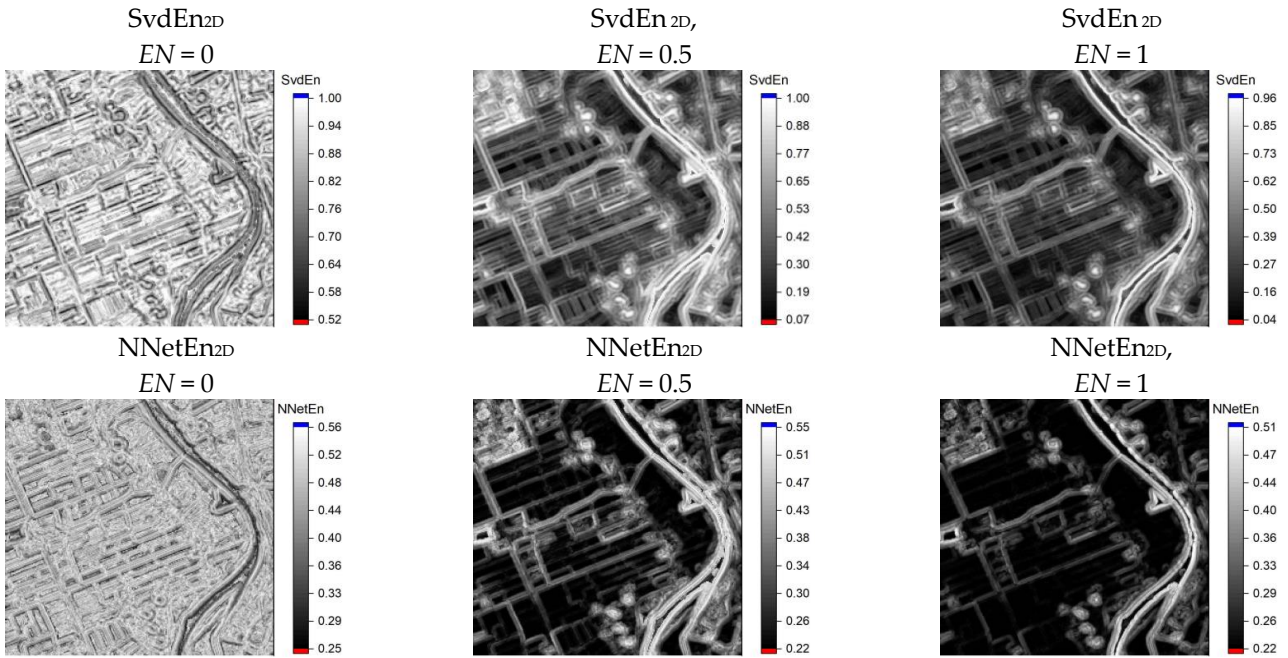

**Figure 10.** Examples of the distribution of SvdEn$_{2D}$ and NNetEn$_{2D}$ for the normalization parameters $EN = 0$, $EN = 0.5$ and $EN = 1$ ($R = 4$).

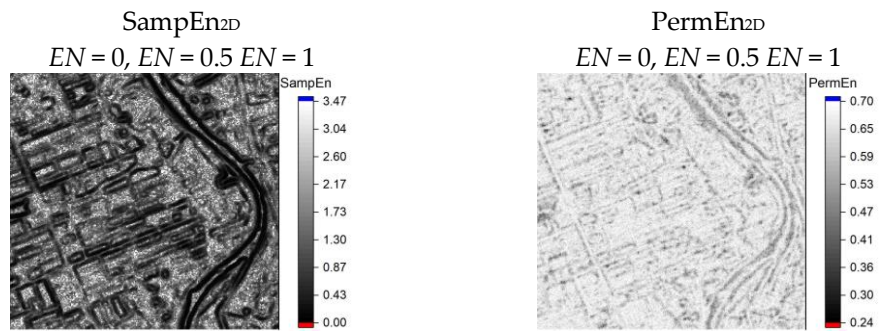

**Figure 11.** Examples of the distribution of SampEn$_{2D}$ and PermEn$_{2D}$ for the normalization parameters $EN = 0$, $EN = 0.5$ and $EN = 1$ ($R = 4$).

### 3.2. Comparison of Regression Algorithms Using Training Set

Comparison of $R^2(1,2)$ metrics for ML regression algorithms using the training set in the SvdEn$_{2D}$ approximation is shown in Table 2. The time for optimizing hyperparameters and training models for each algorithm was fixed at 2 h. Additionally, the table shows the $R^2(172)$ value for image No. 172, which was not present in the training set.

**Table 2.** $R^2$ metrics for ML regression algorithms using training set in SvdEn$_{2D}$ approximation, and $R^2(172)$ values.

| Algorithm | $R^2(1,2)$ | $R^2(172)$ |
|---|---|---|
| gradient boosting | 0.996 | 0.984 |
| support vector regression | 0.982 | 0.891 |
| k-nearest neighbors | 0.982 | 0.889 |
| multi-layer perceptron | 0.972 | 0.864 |
| stochastic gradient descent | 0.970 | 0.848 |
| decision tree | 0.968 | 0.908 |
| automatic relevance determination | 0.872 | 0.840 |
| adaptive boosting | 0.836 | 0.596 |

The best performance (the highest $R^2(1,2)$ and $R^2(172)$) for the SvdEn$_{2D}$ approximation was found for the gradient boosting algorithm. The worst performance for the SvdEn$_{2D}$ approximation was found for the adaptive boosting algorithm. Thus, the $R^2$ metric in Table 2 for training data $R^2(1,2)$ and testing data $R^2(172)$ showed a similar trend across the ML algorithms.

As an example, we present the SvdEn$_{2D}$ profile and the ML_SvdEn$_{2D}$ profiles for the GB and AB algorithms which were calculated from image No. 172 (Figure 12). For the convenience of perceiving the results, we circled all of the 2D entropy distributions that were obtained using ML regression in a brown frame, and we also give the values of $EN$ and $R^2$.

SvdEn$_{2D}$
$EN = 1$

ML_SvdEn$_{2D}$,
$EN = 1$
gradient boosting
$R^2(172) = 0.984$

ML_SvdEn$_{2D}$
$EN = 1$
adaptive boosting
$R^2(172) = 0.596$

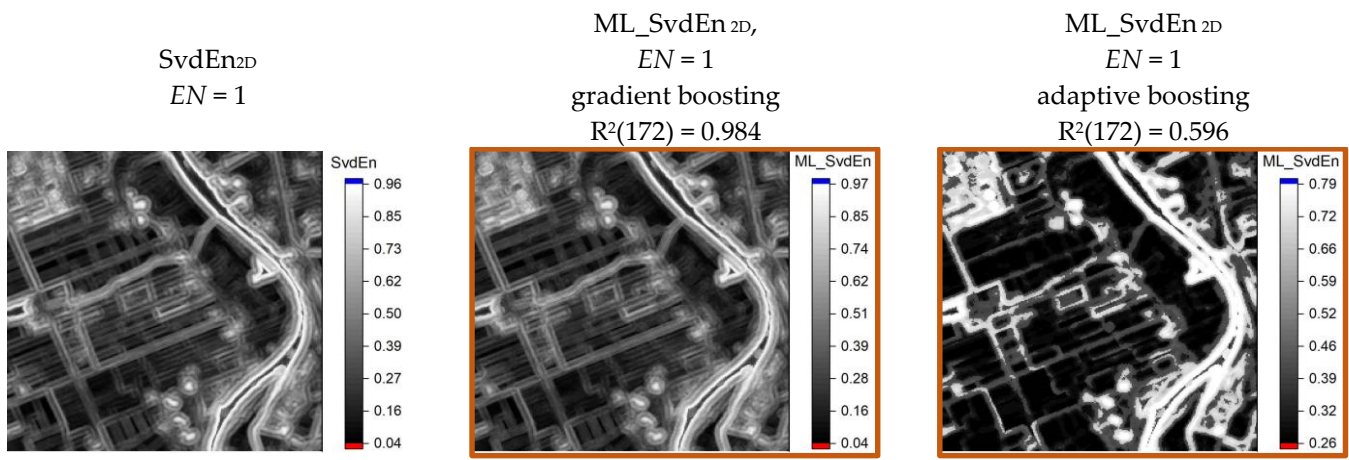

**Figure 12.** SvdEn$_{2D}$ of image No. 172 and ML_SvdEn$_{2D}$ distributions obtained by gradient boosting and adaptive boosting algorithms. Brown frames indicate 2D entropy distributions obtained using ML regression.

It can be seen from the distributions of ML_SvdEn$_{2D}$ in Figure 12 that the gradient boosting algorithm gives a very similar distribution to SvdEn$_{2D}$, in the same range of entropy values from 0.04 to 0.97, while the adaptive boosting algorithm gives a narrower interval from 0.26 to 0.79.

Figure 13 shows a comparison of the profiles for $i = 50$ and the entropies SvdEn$_{2D}$ and ML_SvdEn$_{2D}$ which were obtained by the GB and AB algorithms. It can be seen that the ML_SvdEn$_{2D}$ profile for GB almost coincides with the SvdEn$_{2D}$ profile (red and black lines), while the AB algorithm (blue line) gives significant deviations, especially at low entropy values.

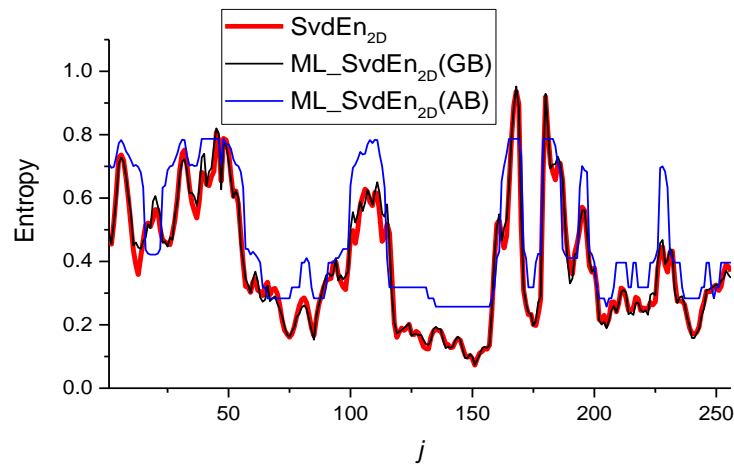

**Figure 13.** Horizontal profile at $i = 50$ for SvdEn$_{2D}$ and ML_SvdEn$_{2D}$ obtained by gradient boosting and adaptive boosting algorithms for Figure 12.

All of the subsequent results are given for the GB algorithm, which showed the best $R^2$ metric.

### 3.3. Results of Approximation SvdEn$_{2D}$ Using GB Regression and Test Set

The value of the $R^2$ metrics (ML_SvdEn$_{2D}$) for all of the 200 Sentinel-2 images (near-infrared band) from the database is shown in Figure 14, where $n$ is the number of the image in the database, and the database includes both training $n = 1, 2$ and testing $n = 3\ldots200$ images. Figure 14 also shows the average levels of $R^2_{mean}$ for two values of the normalization parameters $EN = 0$ and $EN = 1$. The $R^2_{mean}$ values have rather high values $R^2_{mean} > 0.9$, which indicates a good approximation of the GB entropy by regression for all of the images from the database. At the same time, the entropy approximation for $EN = 1$ is, on average, better than it is for $EN = 0$. The standard deviation of $R^2$ for all of the 200 images for $EN = 0$ is $\sigma = 0.021$ and for $EN = 1$, $\sigma = 0.018$.

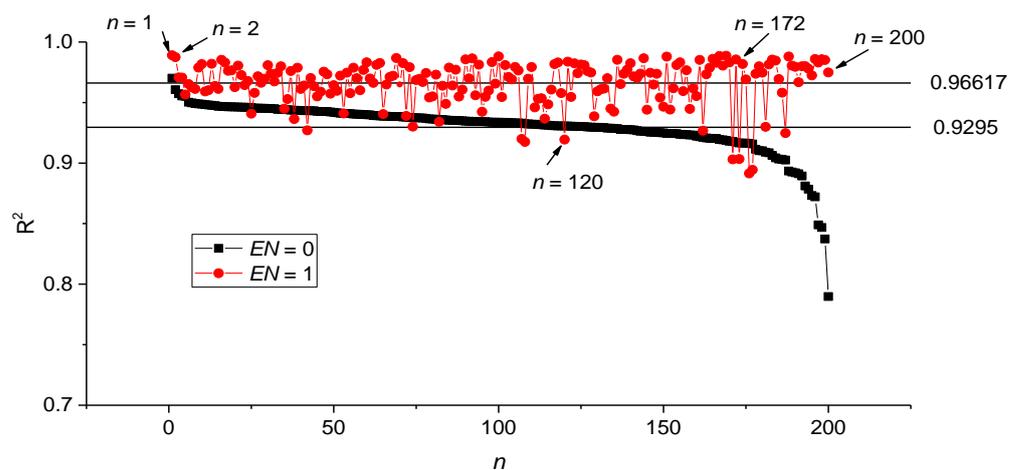

**Figure 14**. The value of the $R^2$ metrics for the Sentinel-2 images database (near-infrared band), for two values of the normalization parameters $EN = 0$ and $EN = 1$ (ML_SvdEn$_{2D}$).

The dependency of $R^2$ metrics on the value of $EN$ for individual images is shown in Figure 15a. In most of the cases, these dependencies have a minimum in the region of $EN$ from 0.1 to 0.3. This is also noticeable in the dependence of the average value of $R^2_{mean}$ on $EN$, which has a minimum of $R^2_{mean} \sim 0.9$ at $EN = 0.1$. The dependence of standard deviation on $EN$ has a maximum value at $EN = 0.2$ (Figure 15b) and a minimum value at $EN = 0.7$.

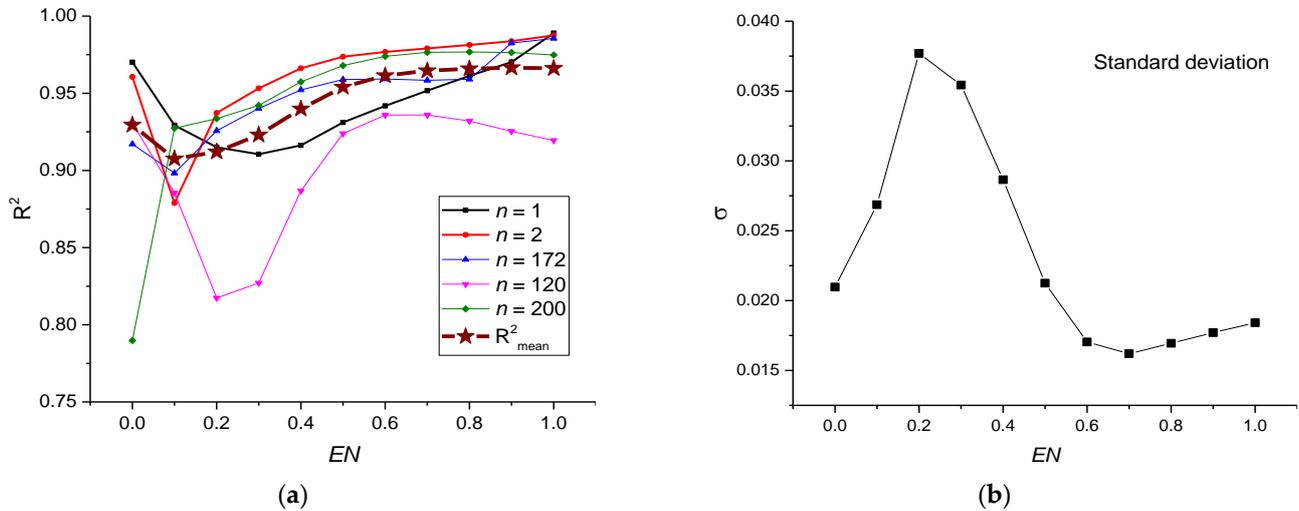

(**a**)                           (**b**)

**Figure 15.** (**a**) Dependence of the $R^2$ metric on the $EN$ value for individual images from the database (No. 1, 2, 172, 120, 200), as well as the dependence of $R^2_{mean}$ on $EN$. (**b**) Dependence of the standard deviation of $R^2$ on the normalization parameter $EN$ (ML_SvdEn2D).

The test results for red, green and blue bands compared to near-infrared bands are shown in Figure 16.

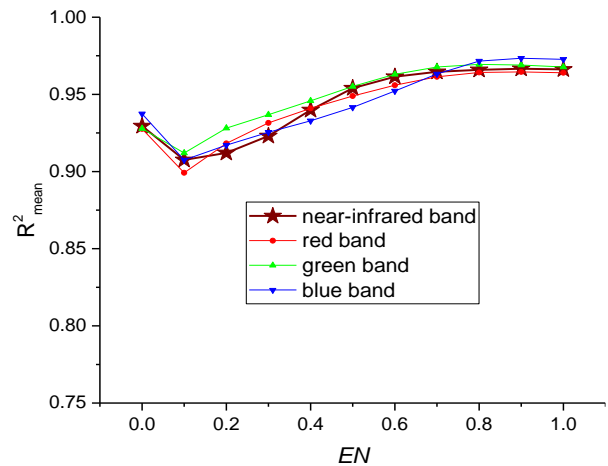

**Figure 16.** $R^2_{mean}$ versus $EN$ for red, green and blue bands versus near-infrared bands (ML_SvdEn2D).

Thus, the trained model approximates the SvdEn value with high accuracy over the entire $EN$ range for the entire set of images in the database for the red, green, blue and near-infrared bands. This indicates a good versatility of the model, which was trained on only two images from the database.

A comparison of the distribution of SvdEn2D and ML_SvdEn2D entropies for image No. 172 is shown in Figure 17, which also shows a comparison of the horizontal profiles of SvdEn2D and ML_SvdEn2D for $i = 50$.

It can be seen that for all of the $EN$ values, the SvdEn2D and ML_SvdEn2D profiles have a similar behavior, with $EN = 1$ being the best profile match that was observed, and the worst match was for $EN = 0.1$, which correlates with the $R^2(172)$ value.

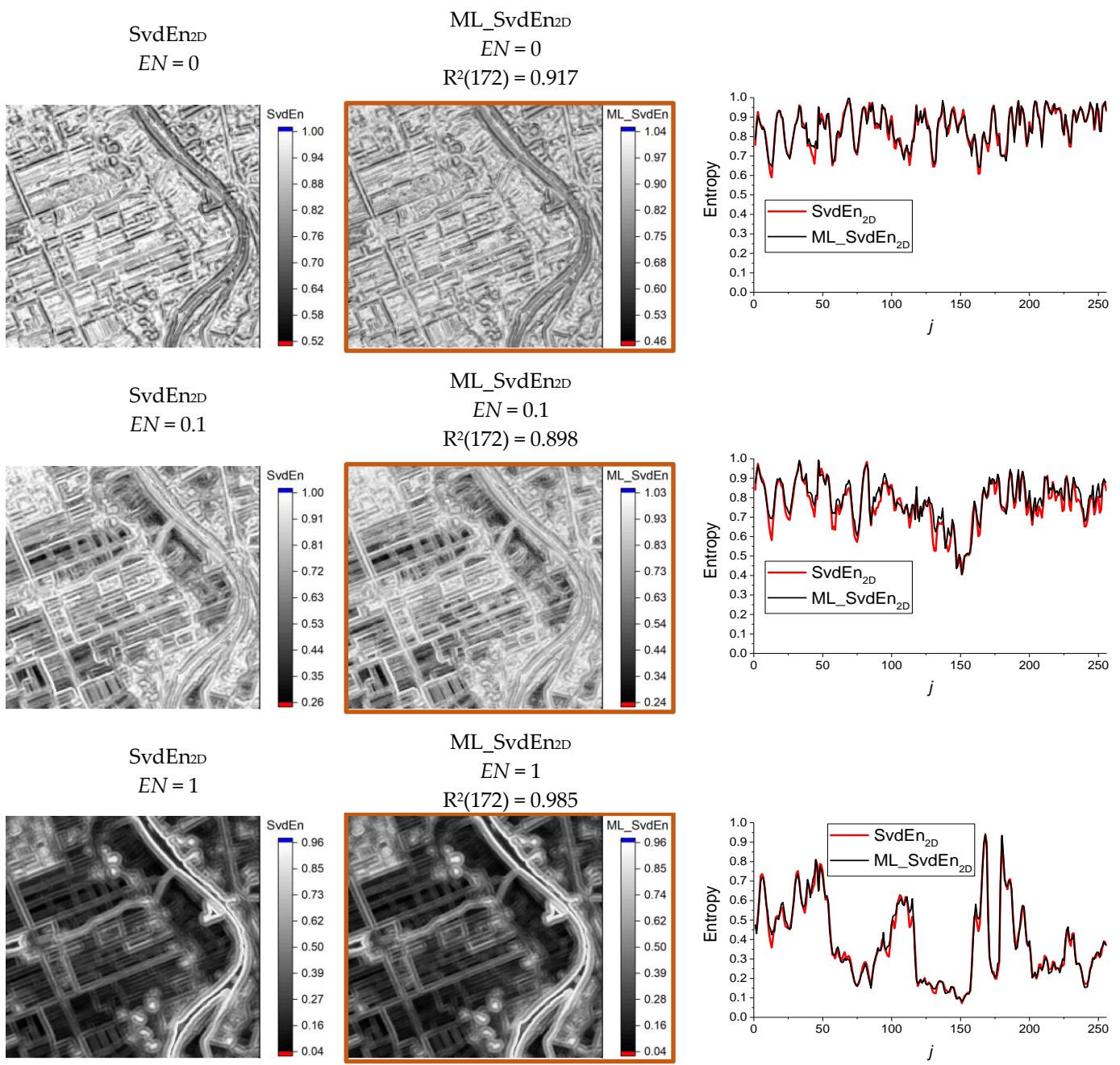

**Figure 17.** Examples of the distribution of SvdEn$_{2D}$ (**left column**) and ML_SvdEn$_{2D}$ (**middle column**) for the normalization parameter $EN = 0$, $EN = 0.5$ and $EN = 1$ ($R = 4$). Comparison of horizontal profiles at $i = 50$ (**right column**).

### 3.4. Results of Fitting SampEn$_{2D}$, PermEn$_{2D}$ and NNetEn$_{2D}$ Entropy Using GB Regression and Test Set

The R$^2$ values of ML_SvdEn$_{2D}$, ML_SampEn$_{2D}$, ML_PermEn$_{2D}$ and ML_NNetEn$_{2D}$ for the normalization parameter $EN = 1$ are shown in Figure 18. The figure also shows the average levels of R$^2_{mean}$. R$^2_{mean}$ have high values of R$^2_{mean} > 0.9$ for ML_SvdEn$_{2D}$ and ML_NNetEn$_{2D}$, and low values for ML_SampEn$_{2D}$ and ML_PermEn$_{2D}$.

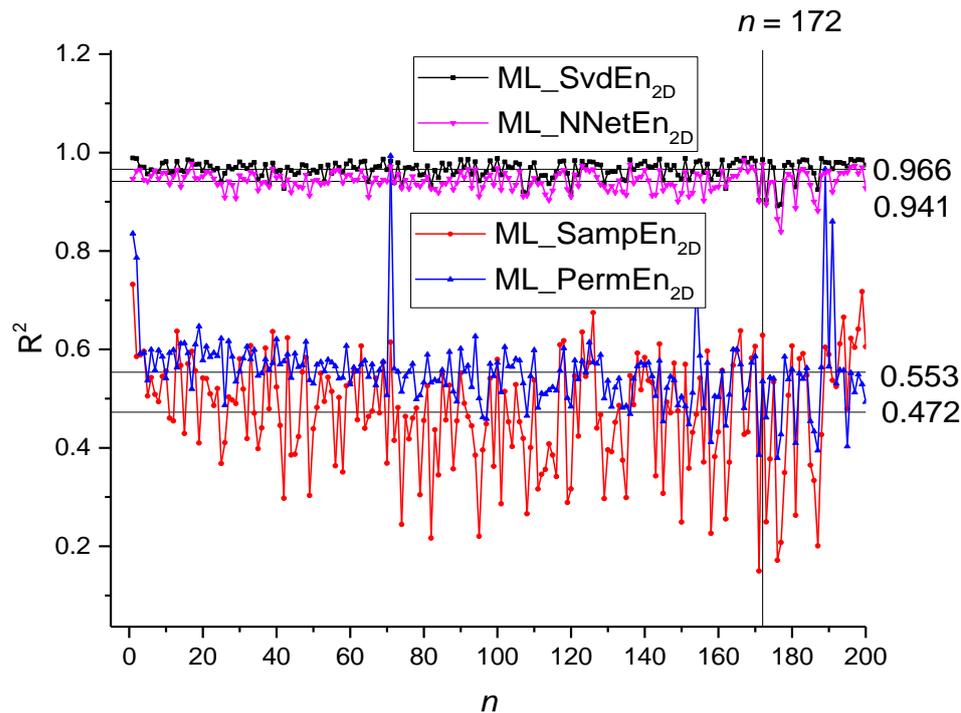

A comparison of the distributions of SampEn₂D and ML_SampEn₂D, as well as PermEn₂D and ML_PermEn₂D for image No. 172, is shown in Figure 19, which also shows the entropy profiles for $i = 50$.

A comparison of the distributions of NNetEn₂D and ML_NNetEn₂D for image No. 172 is shown in Figure 20, which also shows the entropy profiles for $i = 50$.

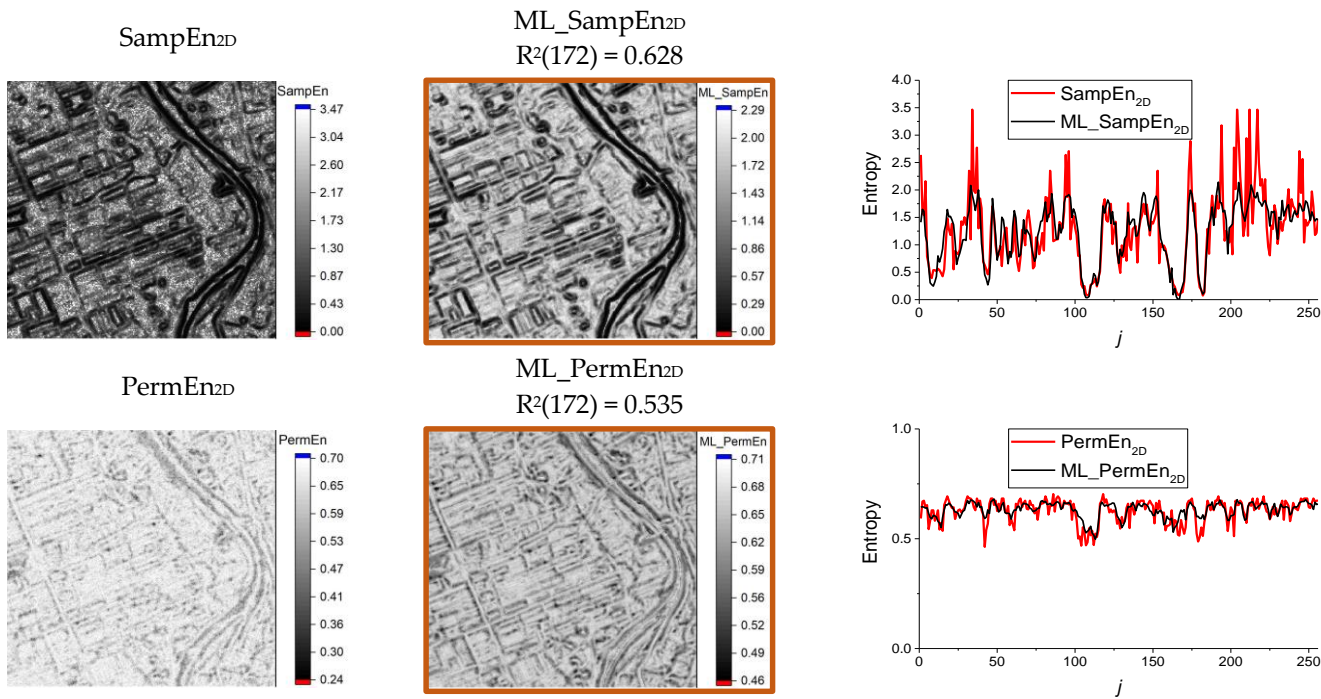

**Figure 19.** Examples of the distribution of SampEn₂D, PermEn₂D and their ML models for the normalization parameter $EN = 1$ ($R = 4$). Comparison of horizontal profiles at $i = 50$ (on the right).

It can be seen that the approximation of SampEn$_{2D}$ and PermEn$_{2D}$ is significantly inferior to the approximation of SvdEn$_{2D}$ and NNetEn$_{2D}$. The distributions and profiles for NNetEn$_{2D}$ practically coincide, which indicates the applicability of the ML regression entropy approximation model. The best distribution contrast of NNetEn$_{2D}$ and ML_NNetEn$_{2D}$ is observed for $EN \geq 0.5$.

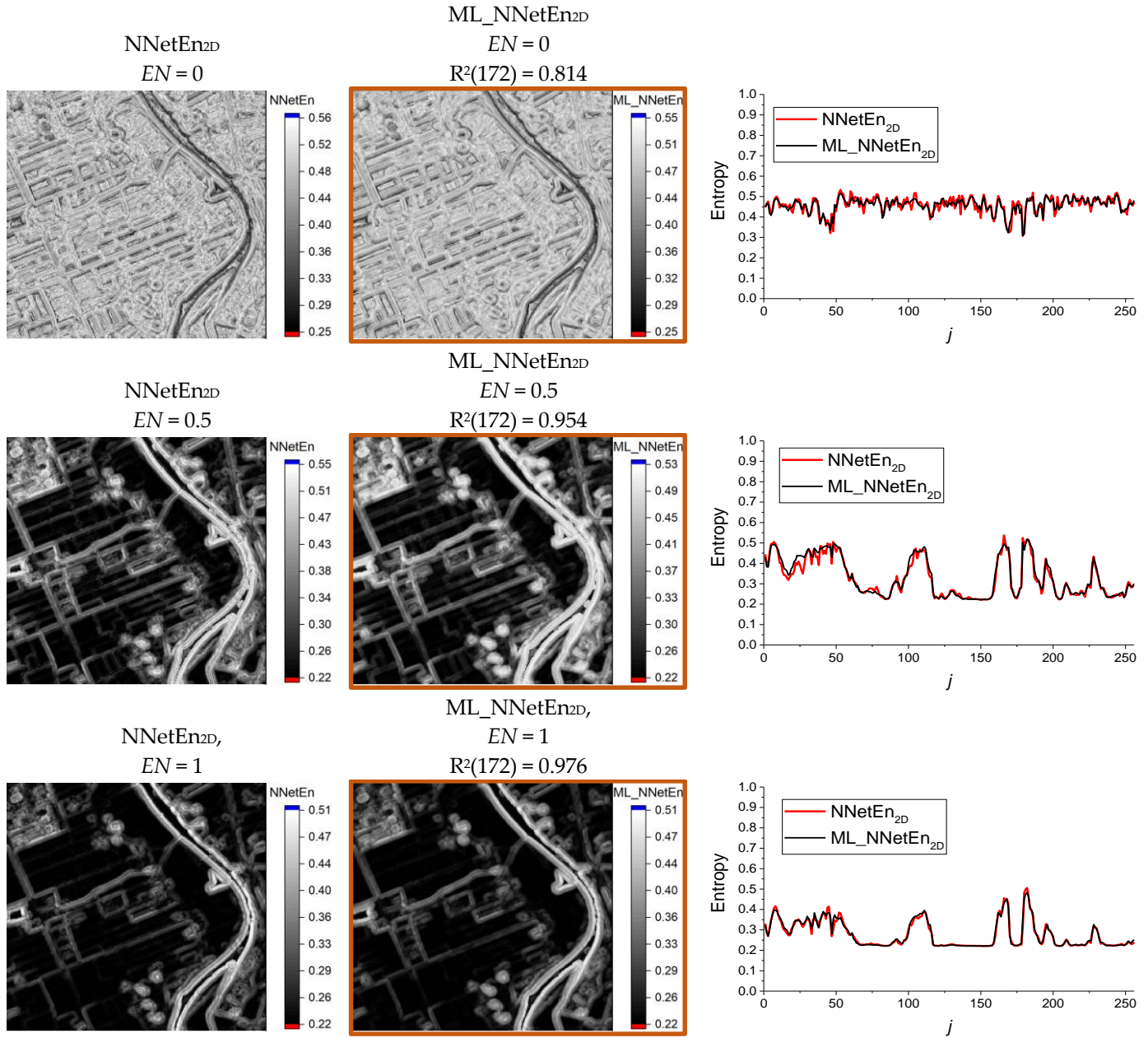

**Figure 20.** Examples of the distribution of NNetEn$_{2D}$ and its ML model (ML_NNetEn$_{2D}$) for the normalization parameter $EN = 1$ ($R = 4$). Comparison of horizontal profiles at $i = 50$ (on the right).

### 3.5. Comparative Characteristics of GB Regression for Approximating Entropies of Various Types and Lengths of the Time Series

Comparative characteristics of the GB regression for the SvdEn$_{2D}$, SampEn$_{2D}$, PermEn$_{2D}$ and NNetEn$_{2D}$ approximation for a spherical kernel with radius $R = 4$ are shown in Table 3. The table shows the $R^2$ metric parameters and time costs when we were calculating one image using one processor thread.

The value of $R^2_{mean}$ has a high value close to 1 for ML_SvdEn and ML_NNetEn, which indicates a good approximation of these types of entropies by the ML model. In addition, the standard deviation values for ML_SvdEn$_{2D}$ and ML_NNetEn$_{2D}$ are low, indicating a good approximation of all of the 200 images from the Sentinel-2 database.

The entropy distribution calculation times for one image with 256 × 256 pixels for SvdEn$_{2D}$, SampEn$_{2D}$ and PermEn$_{2D}$ were calculated in seconds, and they practically coincide with the calculation of their models. There is a slight acceleration of the calculation when we were applying the ML regression for SampEn$_{2D}$ and PermEn$_{2D}$.

The estimated calculation time for one image for NNetEn$_{2D}$ is very long; it is about 8 days, therefore, the calculations were carried out in parallel on 30 processors. At the same time, the ML regression allowed us to calculate ML_NNetEn$_{2D}$ in a couple of seconds, therefore, for this type of entropy, there is an acceleration of the calculation by more than $3 \times 10^5$ times. This acceleration allows ML_NNetEn$_{2D}$ to be widely used in Sentinel-2 image processing.

**Table 3.** Comparative characteristics of GB regression for ML models ($R = 4$, $N = 49$).

| Entropy | ML Model | $R^2_{mean}$ | $\sigma$ | $R^2(172)$ | Entropy Calculation Time, s | ML Model Calculation Time, s | Calculation Acceleration |
|---------|----------|--------------|----------|------------|------------------------------|------------------------------|--------------------------|
| SvdEn$_{2D}$ | ML_SvdEn$_{2D}$ ($EN = 1$) | 0.966 | 0.018 | 0.985 | 2.14 | 3.31 | 0.64 |
| SampEn$_{2D}$ | ML_SampEn$_{2D}$ | 0.472 | 0.113 | 0.628 | 2.35 | 1.26 | 1.86 |
| PermEn$_{2D}$ | ML_PermEn$_{2D}$ | 0.553 | 0.075 | 0.535 | 3.68 | 1.51 | 2.44 |
| NNetEn$_{2D}$ | ML_NNetEn$_{2D}$ ($EN = 1$) | 0.942 | 0.02 | 0.976 | 684195 | 2.01 | 340395 |

Comparative characteristics of the approximation for ML_SvdEn$_{2D}$ at different lengths of time series $N$ using spherical kernels with a radius in the range $R = 1…6$ are shown in Table 4. The best approximation corresponds to the shortest time series $N = 5$ for which $R^2_{mean}$ = 0.997, and it comes close to the maximum value of one. In addition, the standard deviation $\sigma$ for $N = 5$ is abnormally low; it is two orders of magnitude smaller than it is for $N = 113$.

The dependence of $R^2_{mean}$ on the length of the time series $N$ is shown in Figure 21. It can be seen that $R^2_{mean}$ decreases with an increasing $N$ value. Table 4 also shows that the standard deviation for ML_SvdEn$_{2D}$ increases with increasing $N$, and this is reflected in the increase in the difference between the minimum and maximum values of $R^2_{mean}$.

In general, for both $N = 5$ and $N = 113$, there is a good approximation of SvdEn ML by regression with a high value of the metric $R^2_{mean} > 0.9$.

**Table 4.** Comparative characteristics of approximation for ML_SvdEn$_{2D}$ for different lengths of time series $N$ ($EN = 1$).

| $R$ | $N$ | $R^2_{mean}$ | $\sigma$ | $R^2$ Minimum | $R^2$ Maximum | $R^2(172)$ |
|-----|-----|--------------|----------|---------------|---------------|------------|
| 1 | 5 | 0.997 | 0.00094 | 0.99452 | 0.99764 | 0.99881 |
| 2 | 13 | 0.991 | 0.0051 | 0.97063 | 0.99836 | 0.99641 |
| 3 | 29 | 0.977 | 0.012 | 0.92733 | 0.99258 | 0.99129 |
| 4 | 49 | 0.966 | 0.018 | 0.89146 | 0.97029 | 0.98559 |
| 5 | 81 | 0.955 | 0.023 | 0.85312 | 0.99005 | 0.9793 |
| 6 | 113 | 0.947 | 0.026 | 0.82805 | 0.99004 | 0.97186 |

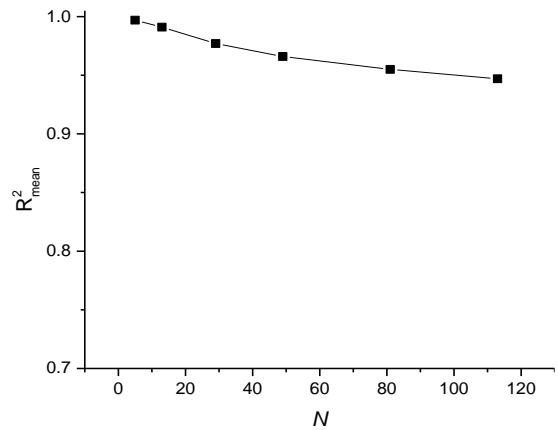

**Figure 21.** Dependence of $R^2_{mean}$ on the length of the time series $N$ for ML_SvdEn2D ($EN = 1$).

Examples of the distribution of ML_SvdEn2D for spherical kernels $R = 1$, 3 and 6 are shown in Figure 22. ML_SvdEn2D for $R = 1$ ($N = 5$) has the sharp boundaries of the regions, while for $R = 6$ ($N = 113$) the entropy distribution pattern looks blurry. A discussion of the advantages of kernels of different radii is given in the discussion.

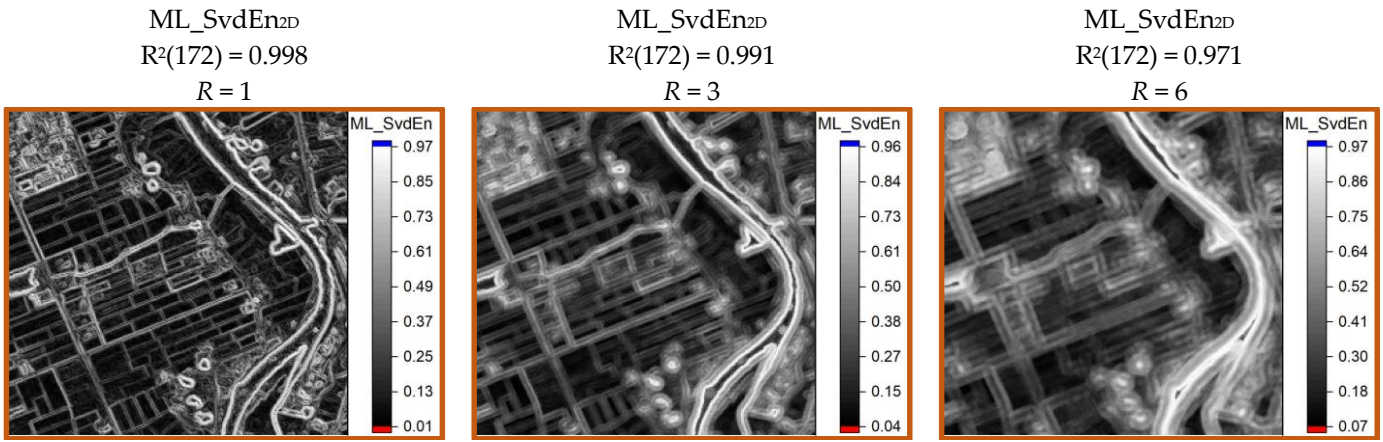

**Figure 22.** Examples of ML_SvdEn2D distribution for spherical kernels $R = 1$, 3, 6 ($EN = 1$).

### 3.6. Results of Approximation of Synthetic Time Series

A bifurcation diagram for logistic map is shown in Figure 23a, and the dependence of SvdEn on the control parameter $r$ in the chaotic time series (Equation (21)) is presented in Figure 23b (black color line). The ML_SvdEn model trained on Plank map is shown in Figure 23b (red color line). It can be noted that ML_SvdEn ($r$) repeated the features of the SvdEn($r$) dependence, and in the range of $r$ = 3.628, 3.742 and 3.838, it also has a minimum value as the original entropy. The largest deviations are observed for the region 3.448 < $r$ < 3.582, where the bifurcation diagram shows the presence of a repeated pattern of several values in the time series. This model has a high Pearson correlation coefficient ~0.968. A model trained on Sentinel images No. 1 and No. 2 is shown in Figure 23c (blue color line), and it has a lower Pearson correlation coefficient ~0.706. The model repeats some of the patterns of the original dependence, and there is a linear shift in the values.

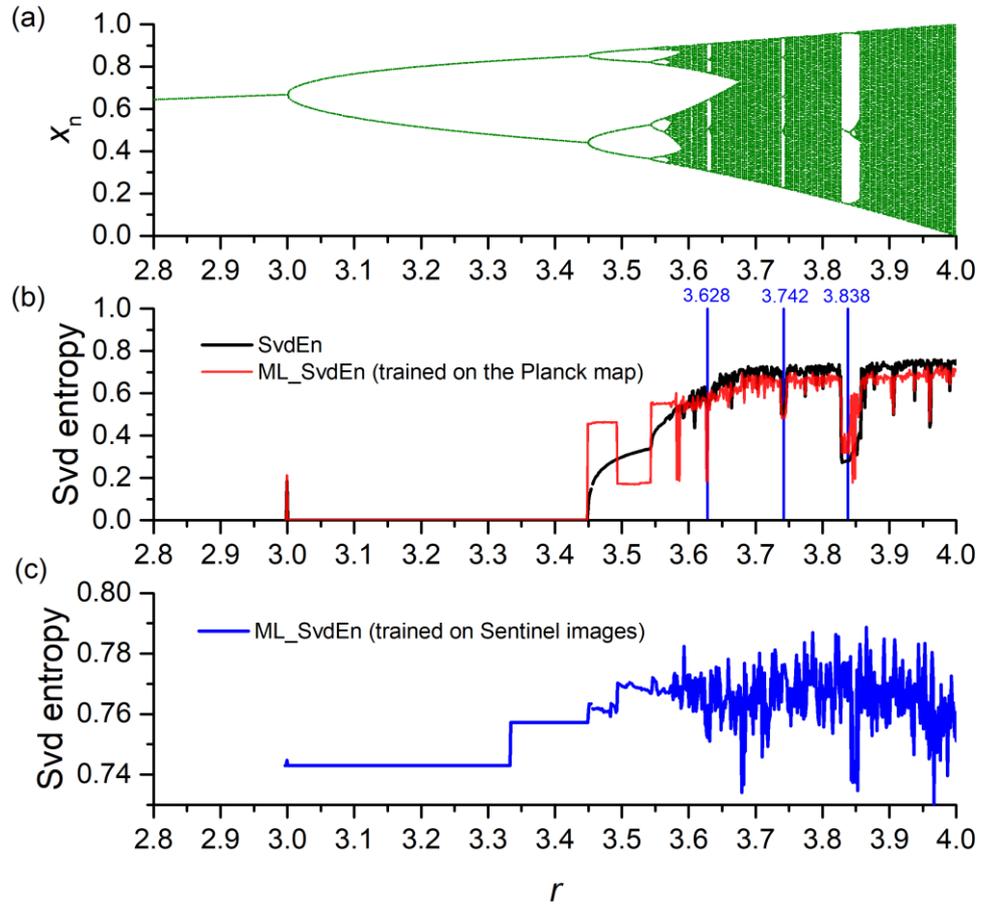

**Figure 23.** Bifurcation diagrams for logistic map (Equation (21)) (**a**); the dependence of SvdEn and ML_SvdEn on the parameter $r$ ($N = 29$, $EN = 0$, $d = 20$ and $delay = 1$) (**b**) the dependence of ML_SvdEn trained on Sentinel images (**c**).

## 4. Discussion

According to the data presented in Figure 18, the ML_SvdEn2D and ML_NNetEn2D regression models approximate the original entropy functions much better than the ML_SampEn2D and ML_PermEn2D ones do. There are three possible reasons for this. Firstly, ML_SvdEn2D and ML_NNetEn2D depend on the $EN$ parameter (see Figure 10); while ML_SampEn2D and ML_PermEn2D are independent of $EN$ (see Figure 11). This leads to the effective size of the training dataset for ML_SampEn2D and ML_PermEn2D being only half the size since it is formed with two images and two values of $EN = 0$ and $EN = 1$. Secondly, it may be related to the discrete nature of the SampEn and PermEn functions. The discreteness of the PermEn values is related to the finite number of permutations (Equation (12)), while SampEn is related to a finite number of template vectors (Equation (14)). In contrast, the calculated SvdEn values are continuous (Equation (10)). Taking into account Equation (15), the NNetEn values can also be considered to be continuous. Regression models are much better at approximating continuous functions, and discrete ones can cause prediction errors. Thirdly, the entropy function is highly nonlinear. For example, Figure 19 shows a section of SampEn2D and ML_SampEn2D, which shows a sharp change in the values of SampEn2D. In the case of an insufficient number of training samples corresponding to the extremes on the curve (Figure 19), the regression model will produce some intermediate values, thereby smoothing the curve and reducing the approximation accuracy. All of the three possible reasons need to be studied in future work.

The obtained results indicate that the ML regression approximates SvdEn2D and NNetEn2D the best ($R^2_{mean} > 0.9$).

A comparison of the sections ML_SvdEn₂D and ML_NNetEn₂D in the area of the image No. 172 is shown in Figure 24. ML_NNetEn₂D more confidently identifies the areas with increased entropy, and it does it a little better than ML_SvdEn₂D. The entropy peaks under labels 2, 3 and 4, which are characteristic of the edges of rivers and reservoirs and are more clearly distinguished in ML_NNetEn₂D. Between labels 2 and 3, the entropy fluctuations are smaller for ML_NNetEn₂D, and this frequency corresponds to cultivated fields, with a relatively smooth texture. The buildings in the region of label 1 has an increased entropy value for both types of entropy. The road labeled 5 also has a sharper entropy peak profile for ML_NNetEn₂D.

The results presented in this paper are of a fundamental nature, as they show the universality of approaches for calculating entropies of various nature, and the possibility of their approximation by the ML regression method.

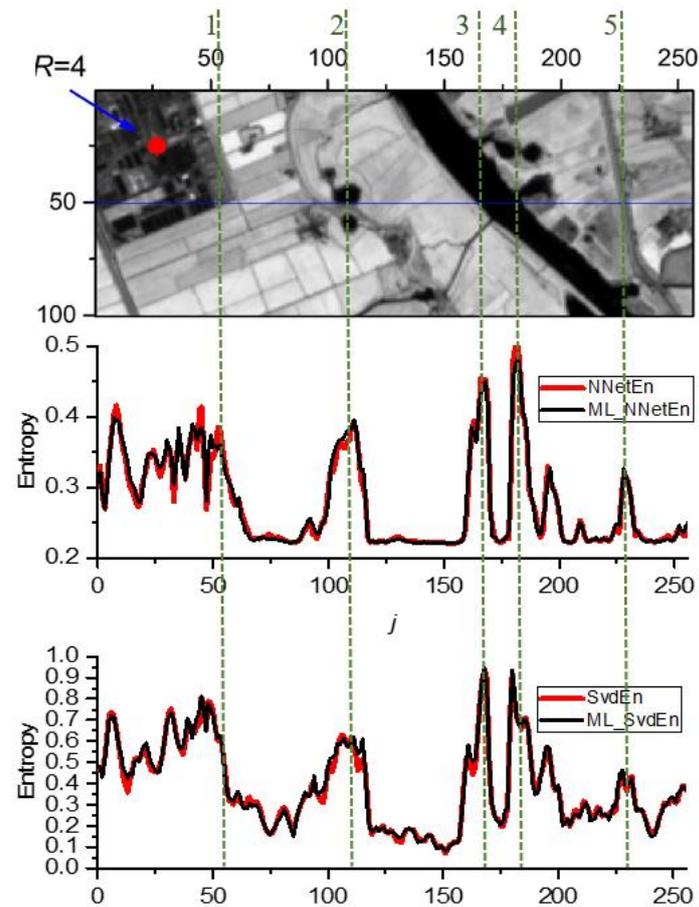

**Figure 24.** Comparison of the sections ML_SvdEn₂D and ML_NNetEn₂D (*EN* = 1, *R* = 4).

From a practical point of view, the entropy calculation using ML models can be conducted faster than the calculation using the original formulas, which is important when one is working with big data. As shown in Table 3, the calculation time can be significantly reduced and an acceleration of several times for PermEn₂D and more than $10^5$ times for NNetEn₂D can be achieved. Due to the fact that NNetEn₂D gives a more informative picture of the entropy distribution than other the entropies can, the acceleration of its calculations is an important achievement.

The paper shows a decrease in the accuracy of the entropy approximation with an increase in the length of the time series (see Figure 21). This trend is understandable, since an increase in the number of elements in the input vector of the regression model leads to a decrease in the accuracy of the regression. Longer time series have more combinations for which the entropy varies significantly. For the short time series with a number of elements of $N = 5$, the regression accuracy reaches $R^2 > 0.99$ and higher, which indicates a very good approximation. An example of an ML_SvdEn$_{2D}$ distribution with a kernel radius value of $R = 1$ ($N = 5$) is shown in Figure 22. The image is distinguished by increased clarity, but it loses its generalizing ability for the cases $R = 3$ and $R = 6$ since entropy is a characteristic of a set of elements. The regression model calculating the entropy for longer time series $N = 113$ had an $R^2 > 0.82$ for ML_SvdEn$_{2D}$, which is also a good result.

In the study, the SvdEn$_{2D}$, PermEn$_{2D}$ and SampEn$_{2D}$ entropies were calculated for the first time using the circular kernel approach. The presented results indicate the applicability of such an approach, and the obtained distributions in Figures 10 and 11 are reflecting the real distribution of the irregularity of the image. The use of circular kernels and ML regression can be applied to other types of entropies in the future.

The results of the approximation of synthetic time series trained on the Planck map showed a high Pearson correlation coefficient of ~0.968 for the logistic map. This experiment showed the versatility of the model, when a model that was trained on one type of data approximates the data of another type well. The most widely used one in the scientific community is the logistic map, for which the regions with high and low entropy which are calculated by different methods are known [33]. The original method SvdEn and its approximation model ML_SvdEn also correctly identify the areas of chaos and order depending on the control parameter $r$. ML_SvdEn ($r$) repeated the features of the SvdEn ($r$) dependence, and in the ranges of $r = 3.628$, $3.742$ and $3.838$, it also has a minimum value as the original entropy (Figure 23b, red line). The largest differences in ML_SvdEn are observed at lower values of SvdEn, where the time series are periodical. We believe that the accuracy of the ML_SvdEn approximation can be increased by improving the time series normalization procedure in further studies. The model trained on Sentinel-2 images has a lower Pearson correlation coefficient of ~ 0.702 for the synthetic time series. However, the values of the high-entropy chaotic series and the local minimum in the range of r = 3.628 and 3.838 were repeated (Figure 23c, blue line). This shows the importance of choosing a training base for the ML entropy approximations. The completeness of the training base plays a decisive role. The model trained on Sentinel images perfectly approximates entropy of the Sentinel images, but it works less well with the synthetic series. The issues related to the optimization of the training base will be considered in subsequent studies.

## 5. Conclusions

The results presented in this paper are of a fundamental nature as they show the universality of the approaches for calculating entropies of various natures (SvdEn, PermEn, SampEn, NNetEn) and the possibility of their approximation by the ML regression method. The high accuracy of the ML models for certain types of entropies are shown: SvdEn$_{2D}$ ($R^2_{mean}$ = 0.966) and NNetEn$_{2D}$ ($R^2_{mean}$ = 0.941). The applicability of the method for short time series with a length from $N = 5$ to $N = 113$ elements is shown. The calculation of the entropy time series with a length of $N \leq 113$ by using the ML regression method can be of great practical use in many scientific and technical fields. A tendency for the $R^2$ metric to decrease with an increase in the length of the time series was found. It is shown that an entropy calculation using ML models can be completed faster than a calculation using the original formulas can, which is important when one is working with big data. The application to remote sensing is shown by calculating the 2D entropy distribution of Sentinel-2 images, and $R^2$ estimates of the approximation error were made. The versatility of the model on a synthetic time series is shown.

and M.B.; resources, M.P.W. and A.T.; data curation, M.B. and M.P.W.; writing—original draft preparation, A.V., M.B., M.P.W. and A.T.; writing—review and editing, A.V., M.B., M.P.W. and A.T.; visualization, A.V. and M.B.; supervision, A.V.; project administration, A.V.; funding acquisition, A.V. and A.T. All authors have read and agreed to the published version of the manuscript.

**Funding:** This research was supported by the Russian Science Foundation (grant no. 22-11-00055, https://rscf.ru/en/project/22-11-00055/, accessed on 22 June 2022)).

**Data Availability Statement:** The data used in this study can be shared with the parties, provided that the article is cited.

**Conflicts of Interest:** The authors declare no conflict of interest.